\documentclass[letterpaper]{article} 
\usepackage{aaai2026}  
\usepackage{times}  
\usepackage{helvet}  
\usepackage{courier}  
\usepackage[hyphens]{url}  
\usepackage{graphicx} 
\usepackage{natbib}  
\setcitestyle{numbers,square,comma,sort&compress}
\usepackage{caption} 
\usepackage{algorithm}
\usepackage{algorithmic}
\usepackage{enumitem}
\usepackage{amsmath}
\usepackage{amssymb}
\usepackage{booktabs}
\usepackage{siunitx}
\usepackage{makecell}   
\usepackage{booktabs,tabularx,siunitx,array}
\newcolumntype{Y}{>{\centering\arraybackslash}p{1.2em}} 
\usepackage{placeins}
\usepackage[T1]{fontenc}
\usepackage[utf8]{inputenc} 
\usepackage{booktabs}
\usepackage{tabularx}
\usepackage{ragged2e}
\usepackage{textcomp} 

\usepackage[edges]{forest}
\usepackage{graphicx} 

\usepackage{array}    
\usepackage{amssymb}  
\newcolumntype{L}[1]{>{\raggedright\arraybackslash}p{#1}}

\usepackage{newfloat}
\usepackage{listings}
\DeclareCaptionStyle{ruled}{labelfont=normalfont,labelsep=colon,strut=off} 
\floatstyle{ruled}
\newfloat{listing}{tb}{lst}{}
\floatname{listing}{Listing}
\usepackage[hidelinks]{hyperref}
\usepackage{hypernat}

\title{Hunt Globally: Wide Search AI Agents for Drug Asset Scouting \\ in Investing, Business Development, and Competitive Intelligence}

\author{
    Vlad Vinogradov\textsuperscript{1},
    Alisa Vinogradova,
    Luba Greenwood\textsuperscript{1,3},
    Ilya Yasny\textsuperscript{1,2},
    Dmitry Kobyzev\textsuperscript{1,2}, \\
    Shoman Kasbekar,
    Kong Nguyen\textsuperscript{1},
    Dmitrii Radkevich\textsuperscript{1},
    Ilya Shkirenko\textsuperscript{1},
    Ivan Izmailov\textsuperscript{1},\\
    Daniil Anisimov\textsuperscript{1},
    Roman Doronin\textsuperscript{1},
    Andrey Doronichev\textsuperscript{1}
}
\affiliations{
    \textsuperscript{\rm 1} Bioptic.io, San Francisco, CA\\
    \textsuperscript{\rm 2} LanceBio Ventures\\
    \textsuperscript{\rm 3} Harvard Business School\\

    info@optic.inc
}

\usepackage{bibentry}

\begin{document}

\maketitle



\makeatletter
\makeatother

\begin{abstract}
    Bio-pharmaceutical innovation has shifted: many new drug assets now originate outside the United States and are disclosed primarily via regional, non-English channels. Recent data suggests over 85\% of patent filings originate outside the U.S., with China accounting for nearly half of the global total; a growing share of scholarly output is also non-U.S. Industry estimates put China at $\sim30\%$ of global drug development, spanning 1,200+ novel candidates. In this high-stakes environment, failing to surface "under-the-radar" assets creates multi-billion-dollar risk for investors and business development teams, making asset scouting a coverage-critical competition where speed and completeness drive value. Yet today's Deep Research AI agents still lag human experts in achieving high-recall discovery across heterogeneous, multilingual sources without hallucinations. We propose a benchmarking methodology for drug asset scouting and a tuned, tree-based self-learning Bioptic Agent aimed at complete, non-hallucinated scouting. We construct a challenging completeness benchmark using a multilingual multi-agent pipeline: complex user queries paired with ground-truth assets that are largely outside U.S.-centric radar. To reflect real deal complexity, we collected screening queries from expert investors, BD, and VC professionals, and used them as priors to conditionally generate benchmark queries. For grading, we use LLM-as-judge evaluation calibrated to expert opinions. On this benchmark, our Bioptic Agent achieves 79.7\% F1 score, outperforming Gemini 3.1 Deep Think (59.2\%), Gemini 3.1 Pro Deep Research (58.6\%), Claude Opus 4.6 (56.2\%), OpenAI GPT-5.2 Pro (46.6\%), Perplexity Deep Research (44.2\%), and Exa Websets (26.9\%). Performance improves steeply with additional compute, supporting the view that more compute yields better results.
\end{abstract}

\begin{links}
\link{Platform}{https://bioptic.io}
\end{links}

\section{Introduction}
Large pharmaceutical companies rely heavily on external innovation to sustain and expand their pipelines, with the majority of new drugs sourced externally through Business Development and Search \& Evaluation (BD and S\&E) functions. While pharmaceutical companies have invested significantly in AI to accelerate internal discovery and development, these efforts represent only a small fraction of pipeline development. In contrast, very little progress has been made in applying AI to drug asset scouting, where a substantial share of strategic value lies. In this high-stakes environment, missing a single qualifying program/asset can mean losing a top partnering or acquisition opportunity worth billions of US dollars, making BD and S\&E competition where speed and completeness matter \cite{schuhmacher2025external,mckinsey2025external}.

\FloatBarrier
\begin{figure}[!t]
  \centering
 \includegraphics[width=1.0\linewidth]{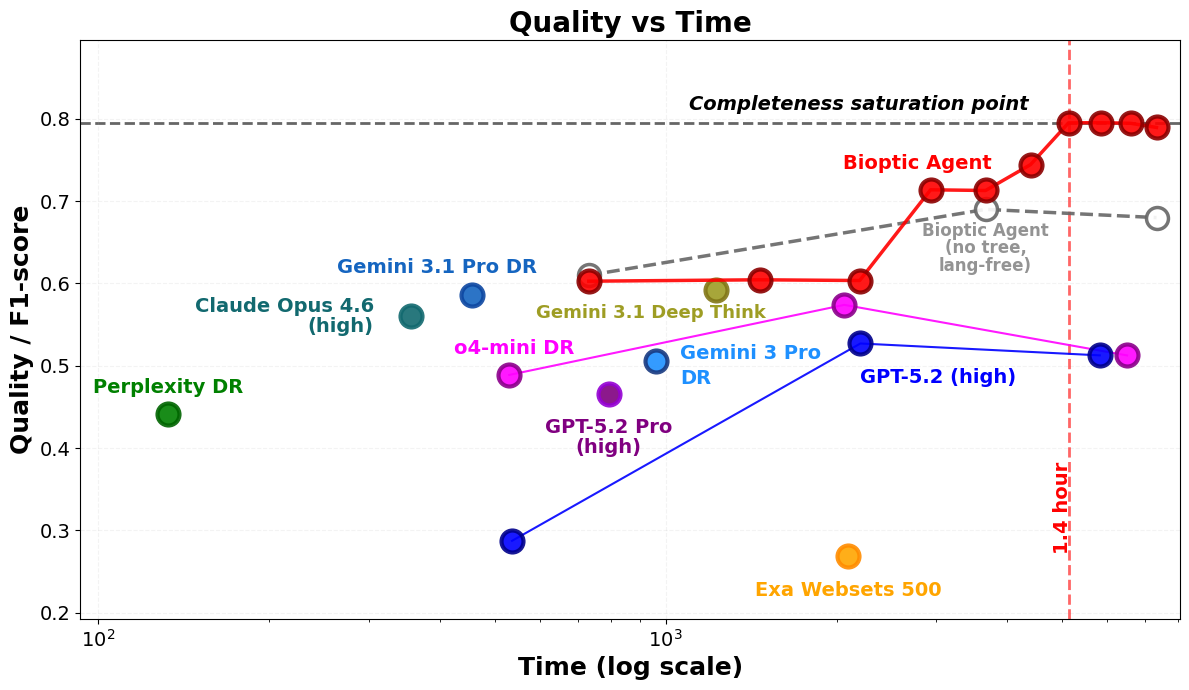}
  \caption{
    \textbf{Quality--time tradeoff for asset scouting.} \emph{y-axis:} F1-score (harmonic mean of precision and recall; higher is better). \emph{x-axis:} wall-clock time (log scale; larger indicates longer compute). \textit{DR} here stands for Deep Research; \textit{lang-free} stands for no language parallelism.
  }
  \label{fig:quality_over_time}
\end{figure}

\begin{figure*}[t]
  \centering
  \includegraphics[width=17
cm]{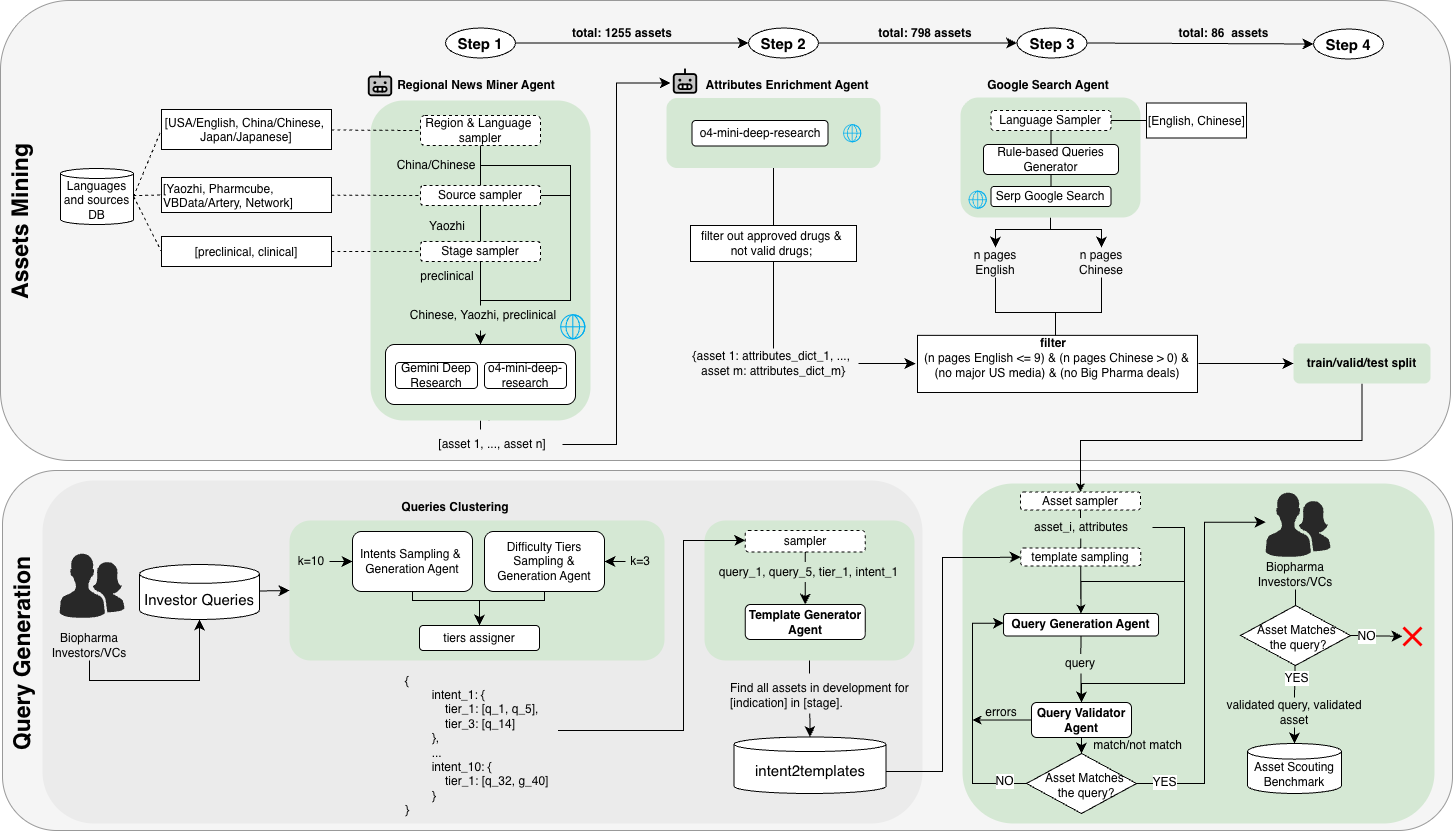}
  \caption[Completeness Benchmark construction pipeline]{%
  \textbf{Completeness Benchmark construction pipeline}
  \textit{Top: Assets Mining} the Regional News Miner Agent surfaces regional-stage drug assets from non-English sources; the Attributes Enrichment Agent validates and structures each asset; the Google Search Agent prioritizes under-the-radar assets via an English vs.\ origin-language discoverability check.
  \textit{Bottom: Query Generation} real Investor Queries are clustered by intent and distilled by the Template Generator Agent into \texttt{intent2templates}; conditioned on these templates, the Query Generation Agent produces benchmark queries paired with ground-truth (GT) assets, and the Query Validator Agent along with human expert validators ensure each query--GT pair is valid and investor-realistic.%
  }
  \label{fig:framework-diagram}
\end{figure*}

Crucially, this scouting problem is increasingly global and often disclosed through regional channels, with a substantial amount of innovation coming from outside of the US. For example, WIPO patent applications by origin (origin data) underscore how widely innovation is distributed. Recent data indicates that \textasciitilde86.5\% of global patent applications originate outside the US, with China being the largest country of origin, accounting for \textasciitilde48.2\% of worldwide applications \cite{wipo_wipi_2025}.
In biopharma, public leadership commentary has similarly emphasized that major portions of development activity now sit outside traditional U.S./EU hubs. For instance, as reported, Pfizer CEO Albert Bourla cited China as \textasciitilde30\% of global drug development with \textasciitilde1{,}200 novel drug candidates \cite{reuters_bourla_china_2025_10_15}. 
Together, these signals motivate a practical shift for scouting where competitive advantage can depend on identifying qualifying programs from heterogeneous, multilingual sources before they are broadly amplified. In such settings, even modest omission rates can be high-cost, which makes completeness and verification central evaluation targets.

As a result, BD and S\&E activities are limited to largely manual and time-consuming tasks, relying on human expertise. To maintain a competitive edge and secure top assets, what is needed are wide-research AI agents for global asset scouting: systems that can systematically identify best-in-class assets across languages and sources faster than rivals and optimize for completeness and match or surpass human expert quality.

Existing deep-research tools are typically optimized for fast web fact-finding and citation-backed report synthesis, and common benchmarks reflect this objective. Benchmarks like BrowseComp \cite{wei2025browsecompsimplechallengingbenchmark} emphasize short browsing tasks with a single verifiable answer, while rubric-based benchmarks such as ResearchRubrics \cite{sharma2025researchrubricsbenchmarkpromptsrubrics} and DRACO \cite{zhong2026dracocrossdomainbenchmarkdeep} evaluate grounding, reasoning, clarity, and citation quality in long-form outputs. Consequently, these evaluation targets favor depth over breadth, where breadth here means completeness-first, open-world set discovery. In BD-style “find-all” searches, this can lead to superficial retrieval and the most novel and valuable programs being missed, including those under the radar or disclosed primarily in non-English ecosystems. DeepSearchQA moves toward exhaustive answer sets, but it typically evaluates tasks where the correct set is smaller, whereas our setting emphasizes long-tail enumeration with larger cardinality, where valid outputs can reach hundreds or thousands of entities \cite{deepsearchqa}. Wide-coverage collection benchmarks further show that exhaustive enumeration is a distinct capability that remains difficult even for strong agents in open-world ``enumerate-all'' settings \cite{widesearch}, and complementary evaluation work cautions that apparent progress on curated tasks can mask persistent omission and shallow-coverage failure modes in realistic browsing and extraction \cite{illusion_of_progress_web_agents}. Hence, current Deep Research agents are still less reliable than human experts at consistently identifying all assets that meet complex, multi-constraint criteria.

One of the reasons why this remains unresolved is that much of the required evidence is public but scattered across heterogeneous, fast-moving, and often non-English sources such as company pages, local press, regional trial registries, conference materials, regulatory and legal filings, and corporate PDFs, which forces scouts to traverse many documents to confirm tight criteria while maintaining provenance and completeness under limited time and budget. The same asset frequently appears under multiple aliases through code-name changes, transliterations, and subsidiary disclosures, and early-stage under-disclosure to protect Intellectual Property (IP) is common, so “what exists” and “what can be found reliably” diverge in real workflows. Some notable data vendors, such as Clarivate and GlobalData, have begun using LLMs to curate drug assets. However, their databases still lack real-time updates and coverage, and they cannot handle complex queries. 
Another point is that the challenge is not only to build an index of all the assets in the world, but also to translate an investor- or diligence-style query into the complete set of matching assets, a step that requires expert interpretation of the technical criteria baked into the query. This intelligence gap is one of the main focuses of our method and evaluation, with precision and coverage as the core metrics.

To make completeness measurable under realistic BD and S\&E conditions, we introduce a completeness-first benchmark for drug asset scouting that is constructed backward from validated program records rather than forward from the query. 
In particular, we mine predominantly non-US regional assets from non-English ecosystems using a multilingual, multi-agent mining configuration that explicitly spans region, language, source type, and development stage. 
We reduce method-induced coverage bias via entity-agnostic query templates and by aggregating candidates across multiple provider deep-research systems. We then validate and enrich candidates into structured program records with provenance, while normalizing aliases and prioritizing under-the-radar discoverability patterns. To ensure the benchmark reflects real screening intent, we ground query construction in expert screening behavior by using BD/VC-style multi-constraint screening queries as priors. This way, success requires evidence aggregation and constraint satisfaction. Because “find-all” evaluation is dominated by entity normalization and attribute verification, the benchmark further operationalizes expert-aligned grading with LLM-as-judge components for alias resolution and up-to-date attribute extraction.

Building on this benchmark, we present Bioptic Agent, a tree-based, self-learning scouting system engineered around completeness and non-hallucination. Instead of compressing exploration into a single evolving narrative, the agent preserves the candidate set and its evidence as persistent artifacts, allocates compute to under-explored branches, and uses expert-aligned critic and validator signals to surface constraint violations and coverage gaps, converting these failure modes into targeted child directives that drive sustained recall growth. This is the key place where general “self-correction” loops remain insufficient in practice, because self-critique can improve internal consistency while still silently failing task-specific constraints without validators tailored to the end screening objective. 

This paper expands our previous work on benchmarking LLM-based agents for competitive landscape mapping in drug asset due diligence \cite{vinogradova2025competitive} by moving from competitor discovery within a single diligence context to open-world, multilingual, “find-all” asset scouting with controlled query intent and difficulty. In a head-to-head evaluation against state-of-the-art commercial deep-research baselines, Bioptic Agent achieves 79.7\% F1-score (harmonic mean of precision and recall; higher is better) and substantially outperforms Claude Opus 4.6 high at 56.2\%, Gemini 3.1 Deep Think at 59.2\%, Gemini 3.1 Pro Deep Research at 58.6\%, OpenAI GPT-5.2 Pro high at 46.6\%, Perplexity Deep Research at 44.2\%, and Exa Websets at 26.9\%, supporting the conclusion that investor-grade and BD-grade scouting requires completeness-oriented search control, lossless candidate tracking, and expert-aligned validation rather than increased browsing compute or better synthesis alone.


\section{Completeness Benchmark}
\label{sec:completeness_pipeline}

Constructing benchmarks for open-world \emph{find-all} asset scouting is prone to method-induced coverage bias: if candidate assets are collected by an agent or expert \emph{given} the query, the resulting ground truth systematically over-represents what that same method can discover. 
To reduce this bias and make the benchmark hard enough, we invert the process. We start with independently observed drug assets curated from regional news sources in local languages. We validate and enrich them with evidence-grounded attributes. We then generate investor-native queries for which the sampled seed asset is a correct answer, while forbidding direct identifiers (names/codes, trial IDs, unique URLs, rare aliases).
Each benchmark sample is a multi-constraint query with a single ground-truth (GT) asset that is intentionally difficult to surface via English-centric search, so success requires evidence aggregation and multi-hop reasoning rather than string matching.


This design may introduce inherent residual bias, that is, seed selection is not uniform across all assets, with news favoring entities with wide media coverage, specific geographies, drug modalities, and certain development stages. Our pipeline mitigates these effects and ensures controlled distribution by (i) filtering out PR noise, (ii) mining across regional news sources using local languages, (iii) filtering out globally amplified assets using an English-vs-local discoverability signal, and (iv) conditioning query generation on a seed corpus of real investor/BD queries to match the empirical intent and constraint-composition distribution, ensuring query realism and GT correctness. Figure~\ref{fig:framework-diagram} represents an overview of the pipeline that creates the benchmark.



\subsection{Regional News Miner Agent}
\label{sec:regional_miner}

To counter the tendency of unconstrained web discovery to overweight English/US announcements and to surface locally announced programs earlier in their public lifecycle, we introduce a Regional News Miner that iterates over tuples \(\langle\)region, language, source, stage\(\rangle\) from a curated region--language--source store (Table~\ref{tab:region_sources}). For each tuple, it executes a deep-research retrieval loop in the \emph{local language} of the target market and returns extracted program (asset) names with canonical links to the underlying announcements.

We construct the store in Table~\ref{tab:region_sources} by manually curating \(10\) regions and \(2\)--\(5\) high-signal biotech news sources per region, prioritizing outlets that routinely publish primary local announcements before they appear in global English coverage. The miner runs a round-robin schedule over the full configuration set \(\mathcal{R}\times\mathcal{L}\times\mathcal{S}(r)\times\mathcal{T}\), where \(\mathcal{R}\) is the set of curated regions, \(\mathcal{L}\) is the set of supported local languages, \(\mathcal{S}(r)\) is the curated set of sources source for region \(r\), and \(\mathcal{T}=\{\text{preclinical},\text{clinical}\}\), ensuring that every \(\langle r,\ell,s,t\rangle\) combination is visited. Specifically, the miner (i) selects the next region--language pair, (ii) selects a curated source within that region, (iii) selects a development stage to control stage coverage, and (iv) runs a search agent that queries two deep-research agents in parallel, OpenAI \textit{o4-mini-deep-research} and Gemini Deep Research, and aggregates their candidates to mitigate provider-specific blind spots. 

The Search Agent constrains Deep Research Agents to query in language $\ell$ and to extract program/asset announcements written in $\ell$ from the designated source $s$. 
Crucially, the search agent is instructed to avoid seeding searches with specific company or drug names and instead use entity-agnostic templates (e.g., new clinical trial authorization, licensing agreement, IND filed, government grant, phase I initiated) combined with the sampled region $r$, local language $\ell$, and source constraints $s$. This reduces popularity and incumbency bias, increases discovery of previously unseen programs, and ensures the benchmark is not anchored to a fixed watchlist. After enumerating all \(\langle\)region, language, source, stage\(\rangle\) combinations, the miner produces \(1255\) validated regional news assets. 
All attributes and supporting artifacts are preserved in the original announcement language $\ell$ and then propagated to the downstream modules of the benchmark pipeline for filtering and validation.

\begin{table*}[!t]
\centering
\small
\setlength{\tabcolsep}{6pt}
\renewcommand{\arraystretch}{1.08}
\begin{tabularx}{\textwidth}{@{}l l >{\RaggedRight\arraybackslash}X@{}}
\toprule
\textbf{Region} & \textbf{Language} & \textbf{Curated sources (2--5 per region)} \\
\midrule
United States & English & FierceBiotech, FiercePharma, Endpoints News \\
China & Chinese & Yaozhi, Pharmcube, VBData/Artery Network \\
Japan & Japanese & Nikkei Biotech, Pharma Japan \\
Korea & Korean & Medigate, ETNews, BioSpectator \\
Brazil & Portuguese & Portal da Saúde, Agência Brasil, Estadão Saúde, Valor Econômico\textendash{}Saúde, JOTA Saúde \\
Australia & English & PharmaDispatch, BiotechDispatch, MTPConnect News \\
Germany & German & Ärzteblatt, transkript.de, CHEManager Life Sciences \\
France & French & Pharmaceutiques, Le Quotidien du Médecin, Biotech Finances \\
Spain & Spanish & Diario Médico, Redacción Médica, El Global \\
CIS countries & Russian/Ukrainian & Vademecum, Pharmvestnik, Apteka.ua \\
\bottomrule
\end{tabularx}
\caption{Manually curated regional sources used by the regional news miner to increase coverage of locally announced programs.}
\label{tab:region_sources}
\end{table*}

\subsection{Attributes Enrichment Agent}
\label{sec:attr_extractor}
Regional News Miner Agent is recall-oriented by design and therefore admits noisy candidates, including false positives and stale metadata (e.g., platform/initiative names misidentified as assets, misspellings and aliases, or outdated information such as ownership and development stage). We therefore validate and enrich with attributes each mined candidate with an \emph{Attributes Enrichment Agent} based on deep-research web retrieval built on top of OpenAI \textit{o4-mini-deep-research}. The agent additionally performs usage-aware cross-lingual entity resolution, retaining the original mention while reconciling local aliases and transliterations and linking them to the commonly used English canonical names observed in global sources.


Given an input asset bundle (the program name as extracted from the announcement, a canonical URL to the announcement page, and the miner context $\langle r,\ell,s,t\rangle$), the agent runs an iterative \emph{search--browse--refine} loop in both the \emph{local language} $\ell$ and English to: (i) verify that the candidate corresponds to a real drug program; (ii) correct misspellings and resolve aliases, including cross-lingual variants and development codes; (iii) determine whether development is currently active; (iv) retrieve and reconcile the most recent development evidence across regions, languages, and source types, preserving multiplicity; (v) extract an up-to-date attributes used downstream for query construction, spanning program-level descriptors and trial-level details, using iterative query expansion to approach evidence saturation; and (vi) find strong global amplification signals (e.g., major US trade-press coverage or large-pharma deal announcements) used to filter out non-challenging for agent's retrieval and over covered in US media assets.

\begin{figure}
  \centering
  \includegraphics[width=1.0\linewidth]{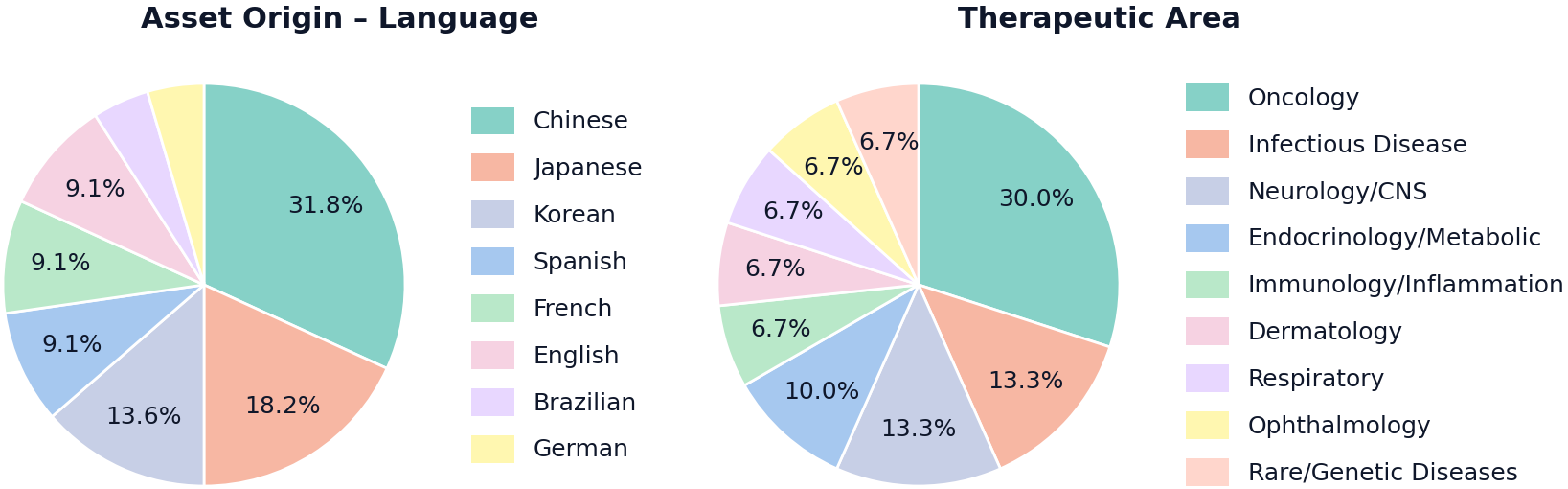}
  \caption{ 
    \textbf{Distribution of asset origin language and therapeutic areas in the benchmark test split.} Left: proportion of assets by origin language. Right: proportion of therapeutic-area labels assigned across assets. 
    }
  \label{fig:language_are_dist}
\end{figure}

The agent is enforced to emit a hierarchically structured attribute record in which every atomic claim is paired with explicit provenance (a list of source URL and verbatim supporting quote pairs). The enriched record contains the following asset attributes:
\begin{itemize}
    \item \textbf{Validation and status:} whether the candidate is a valid drug asset, whether development is currently active, whether it is preclinical or clinical, and the current stage
    \item \textbf{Program descriptors:} developer(s), modality, target(s), short mechanism of action, detailed mechanism of action, indication(s), and patent(s)
    \item \textbf{Trial-level records:} a list of trials, where each trial includes indication, development phase, regimen, efficacy data, safety data, line of therapy, biomarker(s), countries of sites, and endpoints
    \item \textbf{Regulatory fields:} approved geography and regulatory labels
\end{itemize}

Following enrichment, we apply a filtering pass to remove (a) non-drug/invalid/hallucinated entities, (b) globally approved and widely covered drugs (unless approval is local and there is meaningful ongoing development that remains under-amplified in English), and (c) candidates with strong global amplification signals (major US trade press or large-pharma deal announcements). As shown in the pipeline schematic, this step reduces $1255$ mined candidates to $798$ enriched assets, which are propagated to the next module (Google Search Agent).

\subsection{Google Search Agent}
\label{sec:google_scarcity}
For each validated asset we construct a number of queries in English and language $l$, drug's origination country's language.
Using a SERP tool, we count the maximum number of google search result pages for (A) English queries and the max number of pages
for (B) queries in the original local language.
This yields an English-vs-local discoverability profile used in downstream filtering.

\subsection{Conditioned Query Generation}
\label{sec:query_assembling}


Starting from the filtered enriched asset records produced by the Regional News Miner, Attributes Enrichment
Agent and Google Search Agent, we generate queries via an \emph{inverted} procedure (asset $\rightarrow$ query). Given a ground-truth (GT) asset and its validated attributes, we construct a query for which the GT asset is one of the correct answers. We forbid GT identifiers and high-identifiability lexical fingerprints (drug names/codes, trial IDs, asset-unique URLs, rare aliases, and announcement-specific phrasing). We also rewrite constraints (like MoA/target class, modality family, indication with population/line-of-therapy slice, maturity window, geography/ownership, evidence signals) into class-level abstractions where relevant. This ensures that the query returns a candidate cohort that contains the GT asset rather than returning the GT asset as a singleton result. As a result, resolution requires evidence aggregation, multi-hop browsing, and domain reasoning rather than lexical matching.


\begin{figure}
  \centering
  \includegraphics[width=1.0\linewidth]{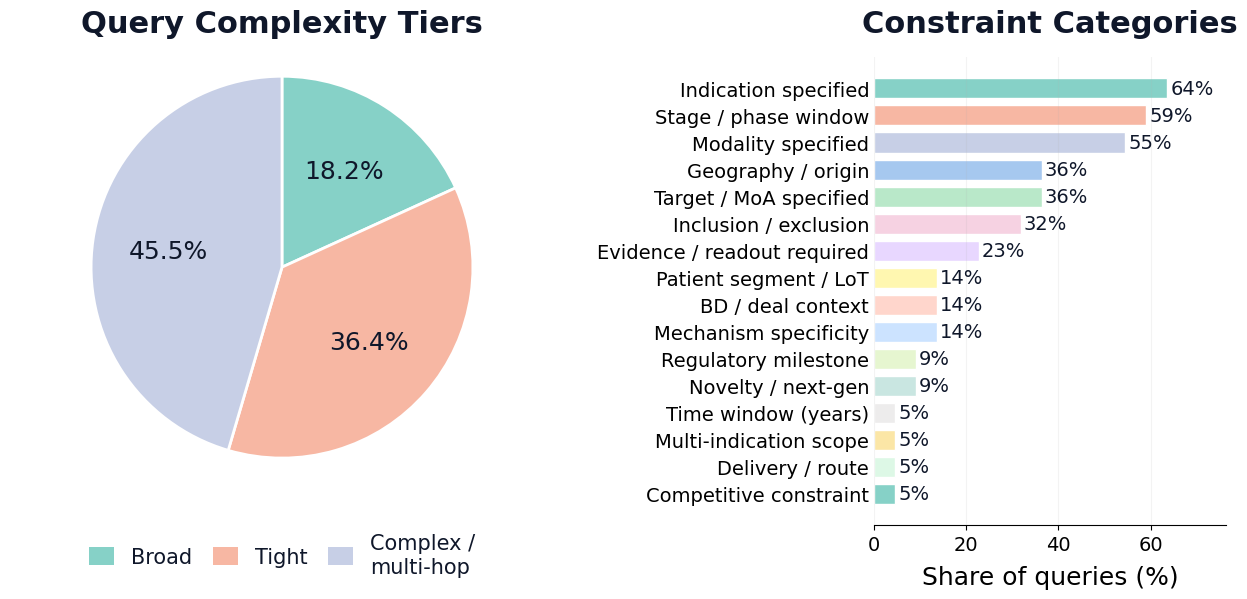}
\caption{
    \textbf{Benchmark query composition.}
    \textbf{Left:} distribution of queries across difficulty tiers (Broad, Tight, Complex/multi-hop).
    \textbf{Right:} prevalence of high-level constraint categories across queries (multi-label).
    A single query can trigger multiple categories; therefore, a category is counted once per query if the query contains that type of constraint.
    }

  \label{fig:query_composition}
\end{figure}

To produce investor-native queries under a constrained generation process, we curate a seed corpus of real biotech investor/BD queries and use it as an empirical prior, ensuring the generated queries match the seed distribution. This corpus consists of $48$ real investor/BD queries that have been reviewed in the coverage pass by expert biopharma investors to ensure the set spans
core diligence workflows. This corpus serves as an empirical prior over (i) \emph{intent} (the business question being asked) and
(ii) \emph{difficulty tier} (how investors combine filters such as modality/platform, target/MoA, indication and population slice,
maturity/stage, geography/origin, ownership/licensing, and evidence signals). Operationally, query generation is conditioned on these corpus-derived dimensions and sampled to match the investor corpus frequency profile, yielding a synthetic query distribution closer to real diligence workloads than unconstrained prompt synthesis.

\textbf{Corpus-derived axes}
From the seed corpus of investor queries, we induce a stable set of intent dimensions using an LLM-assisted clustering procedure. We set the number of intent clusters to $k=10$ to balance granularity and interpretability, and then reassign all $48$ investor queries to these intents. The resulting intent dimensions are:
\begin{itemize}
    \item Program attrition and suspended or terminated programs
    \item Business development screening for in-licensing or acquisition
    \item Indication landscape mapping
    \item Target-first landscape mapping
    \item Precision oncology sub-landscapes
    \item White-space and low-competition target hunting
    \item Geography and origin constraints
    \item Platform and modality scouting
    \item Catalysts and upcoming readouts
    \item Combination regimen opportunity discovery
\end{itemize}

Within each intent, we group the investor queries into three corpus-consistent \emph{difficulty tiers} (see Figure~\ref{fig:query_composition} for their distribution across the final benchmark):
\begin{itemize}
    \item \textit{Broad}: low constraint, primarily enumerative landscape building over a wide search space
    \item \textit{Tight}: conjunctive screening queries with multiple simultaneous constraints on entity class and attributes, designed to filter to a narrow candidate set
    \item \textit{Complex / multi-hop}: tight screening plus at least one derived constraint that typically cannot be validated from a single source and therefore requires integrating evidence across multiple sources, such as lifecycle changes over time, readout or signal aggregation, competitor ceilings (i.e., targeting hot targets; having at most $N$ competing programs), or nuanced include/exclude mechanistic rules.
\end{itemize}

\textbf{Template induction (intent $\times$ difficulty tier) and instantiation.}
For each (intent, difficulty tier) group in the investor corpus, a GPT-5.2 template-induction agent abstracts the group’s real investor queries into a set of structural templates that capture their recurring constraint structure. 
Intent, difficulty tier, and template collectively define a query group used in the conditioned query generation. See Figure~\ref{fig:example_prompt_spec} for representative examples of the query groups used.

\begin{figure}[t]
\caption{Example prompt specifications (intent $\times$ template) used for benchmark query generation. Each \emph{query group} is defined by an intent, difficulty tier, and a template. Here G$i$ denotes the $i$-th query group.
}
\label{fig:example_prompt_spec}
\centering
\small
\setlength{\fboxsep}{6pt}
\setlength{\fboxrule}{0.4pt}
\fbox{%
  \parbox{0.97\linewidth}{%
    \raggedright

    \textbf{[G1]} \par
    \textbf{Intent:} White-space and low-competition target hunting \par
    \textbf{Difficulty tier:} Complex \par
    \textbf{Template:} Find all [stage(s)] [modality] assets targeting [target class] for
    [therapy area/indication], reporting [biomarkers/endpoints/early efficacy/safety] and no more than [N] competitors ahead in [stage(s)];
    exclude tool compounds not intended for patient use. \par

    \vspace{0.6em}\hrule\vspace{0.6em}

    \textbf{[G2]} \par
    \textbf{Intent:} Target-first landscape mapping \par
    \textbf{Difficulty tier:} Complex \par
    \textbf{Template:} Find assets that [target] [target A], across any modality, any indication, and any development phase, focusing exclusively on [target A]
    (do not include multi-target assets unless [target A] is clearly a primary target). \par

    \vspace{0.6em}\hrule\vspace{0.6em}

    \textbf{[G3]} \par
    \textbf{Intent:} Geography and origin constraints \par
    \textbf{Difficulty tier:} Tight \par
    \textbf{Template:} Find [region activity: origin/licensing/trials] [modality] assets [stage] for [disease] targeting the [pathway/axis]. \par

    \vspace{0.6em}\hrule\vspace{0.6em}

    \textbf{[G4]} \par
    \textbf{Intent:} Indication landscape mapping \par
    \textbf{Difficulty tier:} Broad \par
    \textbf{Template:} Find all assets in development for [indication] [stage]. \par
  }%
}
\end{figure}

\textbf{Query Generation}
Given a sampled GT asset, GPT-5.2-based query generation agent (i) selects suitable (intent, difficulty tier) group based on the asset specifics and available attributes (for example if this is the only asset for a specific rare cancer mutation, then it will not pick the group which requires this specifics), (ii) selects a compatible template from that group associated with the difficulty tier, and (iii) generates a query following that template and mapping validated attributes into non-identifying constraints.

\subsection{Query Validator Agent}
\label{sec:validator_editor}
To reduce undetected query–GT inconsistencies and minimize human review load, we run an automated verification loop that flags and corrects likely mismatches before expert inspection. A Criteria Match Validator Agent (the same as the one used in the Bioptic Agent, later discussed in a dedicated subsection of the "Bioptic Agent" section) takes a generated query and a GT asset and predicts whether the asset satisfies the query. 
The validator decomposes the query into atomic criteria with explicit logical
operators (AND/OR/NOT) and comparators, and at each node extracts supporting attributes/evidence
from the structured asset schema.
If the asset does not satisfy the query, the validator returns a failure rationale.
A Query Generator agent (based on gpt-5.2) then revises the query based on the rationale (tightening/loosening
constraints, fixing modality/target/stage mismatches, and removing accidental leakage tokens) returned by the validator.
Revised query–asset pairs are re-validated and revised iteratively until the validator confirms that the asset satisfies the query, after which those pairs are submitted for the expert review.

\subsection{Final filtering and human validation}
\label{sec:final_filtering}
As a final benchmark construction step, we curate query–asset pairs to enforce benchmark hardness while mitigating selection bias toward globally amplified programs.
For the fraction of the assets in the resultant set, we require the asset to have at most $9$ pages of English search results and more than
$0$ pages in the origination language. 
We further validate assets against regional ``white sources'' and conduct final human expert checks and expert query editing. 

For an overview of the final benchmark’s properties, see Figures~\ref{fig:language_are_dist}, \ref{fig:query_composition}, \ref{fig:example_prompt}.
Figure~\ref{fig:language_are_dist} reports coverage by asset origin language and therapeutic area. 
Figure~\ref{fig:query_composition} characterizes query composition by showing both the distribution across constraint tiers and the prevalence of atomic constraint categories. The constraint categories plot is built by tagging each query with the atomic constraint categories, because a query can contain multiple constraints, it can contribute to multiple category bars.
Figure~\ref{fig:example_prompt} shows three randomly sampled queries from the resulting benchmark.




\begin{figure}[t]
\caption{Example queries from the final benchmark, conditionally generated for relevant query groups (G1--G4 in Figure~\ref{fig:example_prompt_spec}).}

\label{fig:example_prompt}
\centering
\small
\setlength{\fboxsep}{6pt}
\setlength{\fboxrule}{0.4pt}
\fbox{%
  \parbox{0.97\linewidth}{%
    \raggedright
    \begin{itemize}
        \item \textbf{[G1]} Find small-molecule antigen presentation modulators in oncology (preclinical through Phase 2). Include only assets with in vivo tumor efficacy and $\leq$ 3 more-advanced direct same-target competitors globally.
        \item \textbf{[G2]} Find all drug assets targeting LAT1, across any modality, any indication, and any development phase, focusing exclusively on LAT1 (do not include multi-target assets unless LAT1 is clearly a primary target).
        \item \textbf{[G3]} Find China-developed/licensed (or have China trial activity) biologic assets (mAbs/bispecifics) in development for autoimmune diseases that modulate the TSHR–IGF-1R axis.
        \item \textbf{[G4]} Find all drug assets currently in preclinical or clinical development for treatment of DMD.
    \end{itemize}
  }%
}
\end{figure}

\section{Bioptic Agent}


Bioptic Agent is designed as a tree-based self-learning agentic system optimized for complete, non-hallucinated biomedical asset scouting. The system addresses the recall stagnation problem common in sequential search agents by combining multi-language parallel investigation, tree-based search ideas generation, and search-quality-weighted rewards. This section provides a detailed technical description of the system architecture, components, and algorithmic mechanisms.


Bioptic Agent employs the following main components:
\begin{itemize}
  \item \textbf{Investigator Agents} execute thousands of web searches to retrieve candidate assets matching the user query, optionally guided by a \textbf{Coach Agent} directive that narrows the search space.

  \item \textbf{Criteria Match Validator Agent} checks each candidate asset against the query criteria and outputs a match verdict plus a detailed, traceable, supported by evidence pass/fail rationale.

  \item \textbf{Deduplication Agent} ensures a global set of validated assets contains only unique assets by removing duplicates and resolving aliases.



  \item \textbf{Coach Agent} generates new exploration directives based on accumulated context: assets found, errors detected, actions performed, queries executed and domains visited so far, etc.
\end{itemize}

The system maintains stores that are regularly updated (at each Rollout, Evaluate and Aggregate steps) and accumulate across epochs:
\begin{enumerate}
  \item $C_{\text{global}}$: Holds all candidate drugs discovered before validation; 
  \item $\tilde{A}_{\text{global}}$: Holds all validated, deduplicated assets;
  \item Evidence stores: all executed web queries - $Q_{\text{global}}$; all visited domains - $D_{\text{global}}$; 
\end{enumerate}

Bioptic Agent treats web exploration as a tree, where each node $n$ stores: (1) $d_n$ - an exploration directive produced by Coach Agent, (2) $\delta_n$ - additional instructions produced by Coach Agent related to the directive
(these are prompt modifications to the Investigator Agent's prompts to explore the directive and override some errors in the initial prompt), (3) $\mathrm{parent}(n)$ - parent node reference, (4) $\mathrm{children}(n)$ - the list of all child node references, (5) $N(n)$ - number of total visits to the node, and (6) $W(n)$ - the total cumulative reward score for the node's directive.

The algorithm runs in a loop for a number of epochs, and for each epoch $e$, the following steps are executed:
\begin{enumerate}
  \item \textbf{Initialize (once):} Create the root node $n_0$ with empty directive $d_{n_0}$ add it to the directive tree, which was empty at this point.

  \item \textbf{Select:} Select $m$ nodes $\{n_i\}_{i=1}^{m}$ via Upper Confidence Bound (UCB) rule (see subsection "Selection") from the tree of directives, where $m$ is the number of parallel explorations per epoch. At the first epoch, when root node node has no children, yet, the root node with an empty directive will be chosen at this step.

  \item \textbf{Rollout:} For each selected node $n$:
  \begin{enumerate}

    \item Instantiate one \textbf{Investigator Agent} per each  language (apply \textit{language parallelism}); In total there will be (number of languages) agents;
    


    \item Provide each agent: (1) the user query, (2) aggregated prior asset findings ($\tilde{A}_{\text{global}}$ and $C_{\text{global}}$ discovered so far), and (3) the node's directive $d_n$ plus directive's additional instructions $\delta_n$.
    \item Each Investigator Agent executes web searches in the specified language and returns candidate drugs; Candidates are then merged across agents/languages.
    \item Update $C_{\text{global}}$ with merged candidates and append executed queries/domains to the evidence stores ($Q_{\text{global}}$, $D_{\text{global}}$).
  \end{enumerate}

  \item \textbf{Evaluate:} For each selected node $n$: 
  \begin{enumerate}
      \item The \textbf{Criteria Match Validator Agent} validates each candidate asset against query criteria, producing (match verdict, pass/fail rationale) per candidate.
      \item Then precision $\text{p}^{(e)}_n$ is computed as the fraction of returned candidates that the \textbf{Criteria Match Validator Agent} marks as valid matches to the query;
      \item Then we run \textbf{Deduplication Agent} on the list of assets that validator agent marked as matching the query; Deduplicator agent resolves duplicates and aliases and returns $\Delta \tilde{A}^{(e)}_n$ a set of valid new unique assets that were not yet discovered before and thus not present in the global assets store $\tilde{A}_{\text{global}}$. 
      \item Then using $\text{p}^{(e)}_n$ and $\Delta \tilde{A}^{(e)}_n$ we compute the \textbf{Node Reward} $r^{(e)}_n$ (see subsection "Evaluate: Node Reward") and update the node's $W(n)$ with it ($W(n)+=r^{(e)}_n$) and also increment the number of visits $N(n)$
      \item Puts failure rationales for assets with match verdict that are in $\Delta \tilde{A}^{(e)}_n$ to the list $F_n$.
  \end{enumerate}
  


  \item \textbf{Backpropagate:} For each node $n$, backpropagate the reward and number of node visits to the root, updating number of total visits to the node $N(\cdot)$ and $W(\cdot)$ along the path (see subsection "Backpropogation").

  \item \textbf{Aggregate:} Merge new unique validated items $\Delta \tilde{A}^{(e)}_n$ across all explored nodes; deduplicate/resolve aliases and append only new unique assets to $\tilde{A}_{\text{global}}$.

  
  \item \textbf{Expand:} If not the last epoch, then for each node $n$:
  \begin{enumerate}

\item \textbf{Coach Expansion:} Conditioned on the current node $(d_n,\delta_n)$, directive lineage (path to root), global memory (e.g., $\tilde{A}_{\text{global}}$, $C_{\text{global}}$), evidence logs $(Q_{\text{global}},D_{\text{global}})$, and the node's validation failures $F_n$ produced by \textbf{Criteria Match Validator Agent} in the current epoch (Step~4), the \textbf{Coach Agent} generates (i) $k$ non-overlapping child directives for the next epoch ${chd_{n}}$ and (ii) $ch\delta_{n}$ per-directive additional instructions / prompt updates to Investigator/Actor prompts.
\item \textbf{Tree Update:} Then Coach agent updates the tree of directives, by adding to the node $n$ as children to the $children(n)$, new nodes initialized by ${chd_{n}}$ as ${d_{n}}$ and $ch\delta_{n}$ as $\delta_{n}$.


  \end{enumerate}
\item \textbf{Repeat} from Step 2.
\end{enumerate}

\subsection{Language Parallelism}

At each epoch, Bioptic runs multiple investigator instances in parallel across languages. The system supports:
\begin{itemize}
    \item \textbf{Language Parallelism}: English plus additional configured languages (e.g., Chinese in our evaluation). Each investigator is instructed to search primarily in its target language, increasing coverage over regional sources, transliterations, and non-English pipeline disclosures.
\end{itemize}


\subsection{Criteria Match Validator Agent}


After Investigator Agents produce the candidate list of assets matching the user query in the Rollout step, those candidate items are filtered by a Criteria Match Validator llm-as-a-judge aligned with domain experts with the 88\% precision as discussed in a subsection "Multi-Agent Debate-Based Tuning". It accepts a single candidate asset and a query and predicts whether the asset matches the query. For each candidate, the validator agent performs hundreds of web searches to perform targeted evidence checks against the query's hard requirements, respecting logical structure (AND/OR) when present. It returns a structured verdict containing:
\begin{itemize}
    \item \textbf{Match verdict}: Boolean indicating whether the candidate matches all criteria.
    \item \textbf{Evidence provenance}: Per-criterion fields containing supporting snippets and source URLs when available.
    \item \textbf{Failure rationale}: A concise reason (confirmation or failure summary) grounded in the evidence fields.
    \item \textbf{Normalized attributes}: Extracted attributes (e.g., phase, modality, indication, MoA, developers) along with citations.
\end{itemize}

Validator outputs are retained for error analysis and summarized by Coach Agent into compact failure patterns that guide subsequent exploration. Also, verdicts it produces are used to compute the local precision score $p^{(e)}_n$ for the node. Also note that, unlike the Precision Grader, that is fixed across all experiments, the Criteria Match Validator prompts may change during tuning because they are part of the search agent rather than the evaluation grader.

\subsection{Deduplication Agent}

Deduplication Agent identifies and removes duplicate assets that may appear under different names, aliases, or synonyms. This step is critical for maintaining data quality in the global asset store $\tilde{A}_{\text{global}}$, as pharmaceutical assets are frequently referred to by multiple identifiers: generic names, brand names, development codes, cross-lingual variants, and historical names that change during development.

The agent accepts $\tilde{A}_{\text{global}}$ and $V^{(e)}_n$, the validated list of candidate assets Investigator Agent returned for the given node after Criteria Match Validator Agent filtering, but before deduplication. The agent operates in two modes, differing in their computational approach and search intensity:

\textbf{Light Deduplication:} This mode processes all $V^{(e)}_n$ items in a single LLM pass or, depending on the size of the full item list, in a few passes by splitting the list into batches, sending each batch to the LLM in one pass, then merging the outputs and passing the merged list to the deduplicator again for a final deduplication pass. 
The agent performs hundreds of web searches to discover aliases, alternative names, development codes, and cross-lingual variants for all items simultaneously, then groups duplicates and selects canonical representatives. In our experiments, light deduplication achieves deduplication quality comparable to the heavy mode in the vast majority of cases.

\textbf{Heavy Deduplication:} This mode processes each candidate item $\in V^{(e)}_n$ sequentially, each in a separate LLM pass, checking it against all existing items in $\tilde{A}_{\text{global}}$ with exhaustive web search verification. The agent performs extensive web searches to discover all possible aliases, previous names, development codes, and cross-lingual variants. It results in hundreds of thousands of searches for large candidate sets, but provides higher confidence in deduplication accuracy. 


In our benchmark evaluations, we use light deduplication as the default, achieving state-of-the-art performance with an F1-score of $79.7\%$. The heavy mode is available for scenarios requiring exhaustive verification.

\subsection{Coach Agent}
\label{sec:coach_agent}

In the \emph{Expand} step, Coach Agent implements the tree: for each node $n$ explored in the current epoch, it proposes $k$ child directives for the next epoch based on the continuously updated memory consisting of the the research history, research gaps present and errors produced by the Investigator Agent when it was computing for the parent's directive, as well as the entire path of directives from the root to the current node, the Investigator Agent's initial instructions. Each child directive consists of a short, high-level description of what angle the Investigator Agent should pursue next, what types of sources to prioritize and additional instructions that specify how Investigator Agent should execute that angle. In effect, $d_n$ focuses the investigators on a particular slice of the original query; 





More specifically, the Coach Agent consumes: (1) the query; (2) the node-path directives (i.e., all directives along the path from the current node to the root); (3) the current node's directive $d_n/\delta_n$; (4) all executed web queries $Q_{\text{global}}$ and visited domains $D_{\text{global}}$ (including those attributable to the current node); (5) $F^{(e)}_n$ - list of failure rationales for all the false positive assets returned by the Criteria Match Validator Agent for the given node; (6) the current $\tilde{A}_{\text{global}}$ (and, when available, $C_{\text{global}}$ to avoid rediscovery); and (7) the Investigator Agent’s initial task prompt.

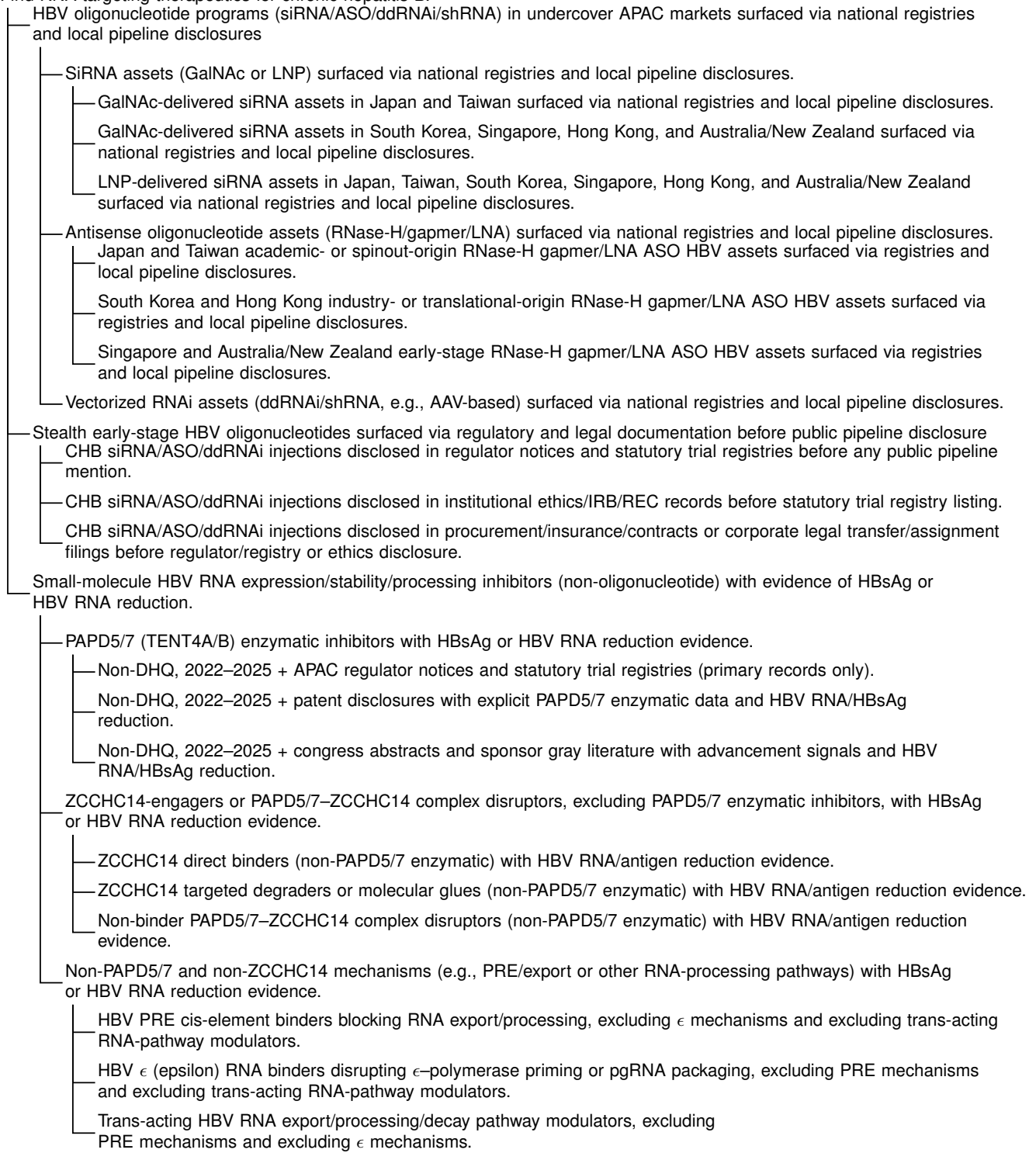
\begin{figure*}[t]
\centering
\resizebox{\textwidth}{!}{%
\begin{forest}
  for tree={
    folder,
    grow'=0,
    fit=band, 
    s sep=0.1cm,
    l sep=0.5cm,
    draw=none,
    edge={draw, semithick},
    anchor=west,
    child anchor=west,
    align=left,
    inner xsep=1pt,
    inner ysep=0.4pt,
    font=\fontsize{7.6}{9}\selectfont\sffamily,
    text width/.expanded={%
      \dimexpr\textwidth-1.5cm-\forestoption{level}\forestoption{l sep}\relax
    },
  }
  [{Find RNA-targeting therapeutics for chronic hepatitis B.}
    [{HBV oligonucleotide programs (siRNA/ASO/ddRNAi/shRNA) in undercover APAC markets surfaced via national registries\\and local pipeline disclosures}
      [{SiRNA assets (GalNAc or LNP) surfaced via national registries and local pipeline disclosures.}
        [{GalNAc-delivered siRNA assets in Japan and Taiwan surfaced via national registries and local pipeline disclosures.}]
        [{GalNAc-delivered siRNA assets in South Korea, Singapore, Hong Kong, and Australia/New Zealand surfaced via\\national registries and local pipeline disclosures.}]
        [{LNP-delivered siRNA assets in Japan, Taiwan, South Korea, Singapore, Hong Kong, and Australia/New Zealand\\surfaced via national registries and local pipeline disclosures.}]
      ]
      [{Antisense oligonucleotide assets (RNase-H/gapmer/LNA) surfaced via national registries and local pipeline disclosures.}
        [{Japan and Taiwan academic- or spinout-origin RNase-H gapmer/LNA ASO HBV assets surfaced via registries and\\local pipeline disclosures.}]
        [{South Korea and Hong Kong industry- or translational-origin RNase-H gapmer/LNA ASO HBV assets surfaced via\\registries and local pipeline disclosures.}]
        [{Singapore and Australia/New Zealand early-stage RNase-H gapmer/LNA ASO HBV assets surfaced via registries\\and local pipeline disclosures.}]
      ]
      [{Vectorized RNAi assets (ddRNAi/shRNA, e.g., AAV-based) surfaced via national registries and local pipeline disclosures.}]
    ]
    [{Stealth early-stage HBV oligonucleotides surfaced via regulatory and legal documentation before public pipeline disclosure}
      [{CHB siRNA/ASO/ddRNAi injections disclosed in regulator notices and statutory trial registries before any public pipeline\\mention.}]
      [{CHB siRNA/ASO/ddRNAi injections disclosed in institutional ethics/IRB/REC records before statutory trial registry listing.}]
      [{CHB siRNA/ASO/ddRNAi injections disclosed in procurement/insurance/contracts or corporate legal transfer/assignment\\filings before regulator/registry or ethics disclosure.}]
    ]
    [{Small-molecule HBV RNA expression/stability/processing inhibitors (non-oligonucleotide) with evidence of HBsAg or\\HBV RNA reduction.}
      [{PAPD5/7 (TENT4A/B) enzymatic inhibitors with HBsAg or HBV RNA reduction evidence.}
        [{Non-DHQ, 2022--2025 + APAC regulator notices and statutory trial registries (primary records only).}]
        [{Non-DHQ, 2022--2025 + patent disclosures with explicit PAPD5/7 enzymatic data and HBV RNA/HBsAg\\reduction.}]
        [{Non-DHQ, 2022--2025 + congress abstracts and sponsor gray literature with advancement signals and HBV\\RNA/HBsAg reduction.}]
      ]
      [{ZCCHC14-engagers or PAPD5/7--ZCCHC14 complex disruptors, excluding PAPD5/7 enzymatic inhibitors, with HBsAg\\or HBV RNA reduction evidence.}
        [{ZCCHC14 direct binders (non-PAPD5/7 enzymatic) with HBV RNA/antigen reduction evidence.}]
        [{ZCCHC14 targeted degraders or molecular glues (non-PAPD5/7 enzymatic) with HBV RNA/antigen reduction evidence.}]
        [{Non-binder PAPD5/7--ZCCHC14 complex disruptors (non-PAPD5/7 enzymatic) with HBV RNA/antigen reduction\\evidence.}]
      ]
      [{Non-PAPD5/7 and non-ZCCHC14 mechanisms (e.g., PRE/export or other RNA-processing pathways) with HBsAg\\or HBV RNA reduction evidence.}
        [{HBV PRE cis-element binders blocking RNA export/processing, excluding $\epsilon$ mechanisms and excluding trans-acting\\RNA-pathway modulators.}]
        [{HBV $\epsilon$ (epsilon) RNA binders disrupting $\epsilon$--polymerase priming or pgRNA packaging, excluding PRE mechanisms\\and excluding trans-acting RNA-pathway modulators.}]
        [{Trans-acting HBV RNA export/processing/decay pathway modulators, excluding\\PRE mechanisms and excluding $\epsilon$ mechanisms.}]
      ]
    ]
  ]
\end{forest}
}
\caption{Part of the Coach Agent generated tree of directives for the query "Find RNA-targeting therapeutics for chronic hepatitis B". Nodes given here are summaries of coach generated directive $d_n$ and directive additional instruction $\delta_n$, we do not include them here in full for readability purposes. It was built for $k=3$. Each node represents an incremental refinement of its parent; the effective search intent at node $n$ is given by the root-to-$n$ path, so child directives omit repeated parent phrasing and specify only the additional constraints or angle.}
\label{fig:scouting_tree}
\end{figure*}

Using this context, then Coach Agent generates:
\begin{itemize}
  \item $d_n$ - child directives for the next epoch, which are (1) non-overlapping concrete refinements (mutations) of the current node's directive, (2) target under-explored search angles suggested by prior outcomes, (3) each strictly narrower than the parent by introducing explicit additional constraints, and (4) collectively exhaustive over the parent’s remaining search space so no major gap is left unassigned.

  \item $\delta_n$ - additional instructions (prompt updates to the Investigator Agent) that reflect observed errors, override ambiguous or incorrect Investigator instructions, and address directive-specific blind spots.
\end{itemize}

The coach is instructed to propose directives that target angles that were missed or only weakly explored, based on the executed-query and domain history ($Q_{\text{global}}$, $D_{\text{global}}$) and recent outcomes, while also exploiting partially explored but promising directions. It uses the compressed validator failure summary (see bellow in \textbf{Failure Summarization}) to refine strategy and reduce recurring rejection reasons in subsequent rollouts. The produced directives must be mutually disjoint (non-overlapping in scope) yet collectively cover the search space defined by the current directive (or the original query when the current directive is empty). Figure \ref{fig:scouting_tree} shows a part of the tree of directives Coach Agent has built for the randomly sampled query from the benchmark.

\textbf{Failure Summarization} To keep its context clean, the Coach Agent calls an LLM to compress $F^{(e)}_n$, the long list of detailed validator failure rationales for all candidates returned by the Investigator agents, into a short set of recurring failure patterns. This is done to mitigate the context rot of the Coach Agent.

\subsection{Evaluate: Node Reward}
This step turns the outcome of the \textbf{Investigator Agent} exploring a directive into the statistics that affect how the \textbf{Selection} step will look at each node. So, here we update the node associated with the directive with the score related to the directive's performance. 



The node reward is computed as:
\begin{equation}
r^{(e)}_n = p^{(e)}_n \, \big|\Delta \tilde{A}^{(e)}_n\big|.
\end{equation}
Here $p^{(e)}_n$ is the local precision for node $n$ at epoch $e$, the fraction of candidates returned by \textbf{Investigator Agent} that the \textbf{Criteria Match Validator Agent} marks as valid matches to the query; $\Delta \tilde{A}^{(e)}_n$ denotes the set of newly added validated and deduplicated assets credited to node $n$ at epoch $e$. These are assets that were not in the global asset set $\tilde{A}_{\text{global}}$ before epoch $e$ and were added due to the rollout under directive $d_n$. 

This reward is designed around the asset scouting objective. We want to prioritize directives that reliably add new valid assets to the global results, rather than directives that only produce many candidates. The precision gate enforces this by reducing the credit given to low-quality rollouts, so high-volume but noisy directives do not attract more compute. 


\subsection{Backpropagation}
\label{sec:backprop}

After computing $r^{(e)}_n$, we backpropagate it by updating node statistics for node $n$ and all of its ancestors in the directive tree:
\begin{align}
N(k) &\leftarrow N(k) + 1, \\
W(k) &\leftarrow W(k) + r^{(e)}_n, \qquad \forall k \in A_n,
\end{align}
where $A_n$ is the set of ancestors of node $n$ including $n$ itself. The resulting values $N(\cdot)$ and $W(\cdot)$ are the quantities used by the UCB rule in \textbf{Selection} step, so directives that repeatedly add new validated assets receive higher scores in future epochs.

\subsection{Selection}
\label{sec:ucb-selection}

At the selection stage, our goal is to decide which search directive to try next so that we keep discovering new valid assets rather than repeatedly visiting the same sources. The Coach agent organizes search into a tree of directives, where each node corresponds to one directive and edges represent refinements proposed during expansion. At any point, the eligible choices are the current leaf nodes, meaning directives that have been generated but do not yet have child directives. In each epoch, we select up to $m$ ($m$ - maximum number of parallel explorations) eligible leaf nodes for parallel rollout, where the Investigator agents execute web searches guided by the selected directive.

We rank eligible leaf nodes using an Upper Confidence Bound rule:
\begin{equation}
\mathrm{UCB}(n) = \frac{W(n)}{N(n)} + c \sqrt{\frac{\log\!\big(\max(1, N(\mathrm{parent}(n)))\big)}{N(n)}},
\end{equation}
where $c = 1.2$ is the exploration constant, $N(n)$ is the number of times node $n$ has been selected, $N(\mathrm{parent}(n))$ is the parent’s visit count and $W(n)$ is the cumulative reward attributed to node $n$ over past epochs. The first term is the empirical mean reward of the directive, that prioritizes nodes whose directives have historically produced higher reward, and the second term is an exploration bonus that prioritizes nodes that have been selected fewer times. Any node that has never been selected receives an infinite score $\mathrm{UCB}(n)=+\infty$, which guarantees it will be tried at least once.



Selection proceeds top-down from the root of the directive tree produced by the Coach Agent. At each internal node, we select the child with the highest UCB score and repeat until reaching a leaf node. When multiple children have the same visit count, we break ties by preferring the child with the higher mean reward; if ties still remain, we break them arbitrarily. This traversal implicitly prioritizes promising branches: directives deeper in the subtree of a high-scoring child are promoted earlier in our evaluation queue, so they are rolled out, scored, aggregated, backpropagated, and expanded sooner than lower-scoring alternatives.

\section{Experiments}
\label{sec:experiments}

We evaluate whether a \emph{specialized} scouting scaffold (Bioptic Agent) outperforms general-purpose generic deep research / find-all systems on the task of asset scouting.
All methods are evaluated on a held-out gold test split consisting of 22 pairs of query-asset from the \emph{Completeness Benchmark} (see Figures~\ref{fig:language_are_dist}, \ref{fig:query_composition} for distribution) using recall, precision, and F1-score.



\subsection{Precision \& Recall Graders}

To evaluate agent performance on the completeness benchmark, we employ LLM-based graders that assess precision and recall against ground-truth assets, in a similar way we did in \cite{vinogradova2025competitive}. Both graders use GPT-5.1 with temperature $0.2$ and a fixed seed for reproducibility, and are equipped with web search tools to verify drug attributes and resolve aliases.

Both graders are designed to be evidence-grounded: all decisions are supported by verbatim quotes and source URLs from authoritative sources, ensuring reproducibility and auditability of the evaluation process.

\textbf{Recall Grader:} The recall grader determines whether the expected ground-truth asset is present in the agent's predicted list of assets. Given the query, the predicted asset list, and the expected asset's attributes (drug name, English name, indication, mechanism of action, modality, developers, stage, country), the grader performs web searches to discover all aliases, alternative names, and cross-lingual variants for both the expected asset and each predicted asset. It then determines equivalence by matching attributes and aliases, accounting for naming variations such as brand names, generic names, development codes, and synonyms. The grader returns a binary verdict: the expected asset is present in the predicted list if any predicted asset matches the expected asset after alias resolution, and absent otherwise.




For each benchmark query--asset example $i$, the Recall Grader outputs a binary verdict $R_i \in \{0,1\}$ indicating whether the expected asset is present in the agent's predicted list after alias/cross-lingual resolution. The reported recall is the mean of these verdicts over the benchmark:
\[
\text{Recall} = \frac{1}{N_{\text{bench}}}\sum_{i=1}^{N_{\text{bench}}} R_i,
\]
where $N_{\text{bench}}$ is the number of benchmark query--asset examples in the evaluated split.

\textbf{Precision Grader:} The precision grader validates each predicted asset against the search criteria to determine whether it should be included in the results. The grader employs a structured decomposition approach: it breaks the query into logical components, then further decomposes each dimension into atomic components. For each atomic component, the grader forms a verification query, performs web searches to retrieve the asset's corresponding attribute value, and documents all steps taken to retrieve this value with verbatim quotes and source URLs. The grader then produces a verdict for each dimension indicating whether the asset meets the criterion defined in that dimension. Finally, the grader assembles all dimension-level verdicts, preserving the query's logical operators (AND/OR), strictness, and overall logic, to produce a final answer of whether the asset matches the query, supported by detailed justification text.


The Precision Grader is applied to every predicted $(\text{query}, \text{asset})$ pair produced by the agent. Let $A_{\text{all predicted}}$ be the set of all such predicted pairs across the benchmark, and let $A_{\text{correctly predicted}} \subseteq A_{\text{all predicted}}$ be the subset that the Precision Grader validates as \emph{match} (i.e., satisfies the query constraints). We report precision as the fraction of predicted assets that are validated as matches:
\[
\text{Precision} = \frac{|A_{\text{correctly predicted}}|}{|A_{\text{all predicted}}|}.
\]

\textbf{Multi-Agent Debate-Based Precision Grader Tuning} The Precision Grader was tuned through an iterative refinement process involving multi-agent debate and expert alignment. Each refinement iteration involves the following components and steps:
\begin{itemize}
    \item \textbf{Generator Agent:} given a query, generate a list of candidate drugs that may satisfy the query constraints. The Bioptic Agent is an example of a generator agent.
    \item \textbf{Precision Grader (tuned):} for each $(\text{query}, \text{drug})$ pair, output a match/non-match verdict with structured justification and provenance.
    \item \textbf{Critic Agent:} evaluate the Precision Grader’s justification, flag disagreements and ambiguous cases, and challenge inconsistent decisions.
    \item \textbf{Debate step:} the Precision Grader and the Critic Agent exchange structured reasons (with provenance) until they converge on a consensus match/non-match verdict (a pseudo-label) for each $(\text{query}, \text{drug})$ pair.
    \item \textbf{Prompt-fix step:} conditioned on the Precision Grader prompt and the post-debate pseudo-labels, the Critic Agent produces a structured summary of \textbf{prompt fixes} explaining the root cause of mismatches; these fixes are manually curated and applied to update the Precision Grader prompt before the next iteration.
\end{itemize}

Because we did not have an initial expert-labeled set large enough to support prompt tuning, we employed a debate-based \textbf{weak-supervision} in which the Precision Grader and the Critic Agent converge on pseudo-labels for each $(\text{query}, \text{drug})$ pair. We operationalize each $(\text{query}, \text{drug})$ as a binary classification instance with the positive class defined as match, i.e., the drug satisfies the query constraints. The system also checks for cross-policy consistency, ensuring the Precision Grader applies consistent criteria for previously identified matches and non-matches.

On a calibration subset of 57 query-drug pairs, the grader achieved $87\%$ accuracy, $100\%$ precision, $82.8\%$ recall, and $90.6\%$ F1-score with respect to the \textbf{debate-converged pseudo-labels} (internal Precision Grader and Critic Agent agreement). After these pseudo-metrics stabilized, domain \textbf{experts performed a one-time audit} of the grader’s predicted match pairs to estimate the trustworthiness of positive decisions in practice; this yielded $88\%$ precision under expert labels. Overall, this design concentrates scarce expert time on the most error-prone cases while still providing a human-grounded calibration of the final grader’s positive predictions.

\subsection{Models \& Agents Tested}
We compare Bioptic Agent against State-of-the-Art search agents:
\begin{enumerate}
    \item \textbf{Find-all agents:} \textit{Exa Websets} ($num\_matches=500$)
    \item \textbf{Deep research agents:} systems optimized for prolonged browsing + synthesis (e.g., \textit{Gemini 3.1 Pro Deep Research}, \textit{Gemini 3 Pro Deep Research}, \textit{Perplexity Deep Research}).
    \item \textbf{Powerful single-pass model:} \textit{GPT-5.2 Pro} in a high search context and reasoning regimes; \textbf{Claude Opus 4.6} in a high search context regime and with 1M tokens context window enabled for the fraction of the benchmark, for which 200k tokens context window was not enough.
    \item \textbf{Sequential scaffold ablations:} \textit{o4-mini-deep-research} and \textit{GPT-5.2} run as a simple iterative loop:
    given previously-found assets, prompt to find more, append new candidates, repeat.
    \item \textbf{Scientific Research \& Reasoning Agents:} \textit{Gemini 3.1 Deep Think} evaluated through their publicly available UI app.

    \item \textbf{Bioptic Agent (no-tree, lang-free) ablation:} removes tree-structured exploration and disables language parallelism, while keeping the remaining components unchanged.
\end{enumerate}

\subsection{Experimental Settings}

Unless stated otherwise, all agents from above are used in their highest supported compute setting, and same prompts (main task prompt and the previous findings prompt that pre-prompts previous found assets) as that was used to initialize the Investigator in the Bioptic Agent are used for all the agents for fair comparison. 

\textbf{Sequential scaffold ablations}
\textit{o4-mini-deep-research} and \textit{GPT-5.2} are wrapped in a vanilla sequential-epoch ensemble. Epoch 1 runs the raw query; each subsequent epoch appends a pre-promted all previously found assets (at previous epochs) and explicitly instructs the agent to return only new items not already listed.

\textbf{Bioptic Agent}
In Figure~1, as wall-clock time increases, the Bioptic Agent explores deeper levels of the directive tree. For all experiments reported in Table~2 and Figure~1, we used \textit{gpt-5.2} with high search context and high reasoning effort as the LLM for the \textit{Investigator Agents}; \textit{gpt-5-mini} with medium search context and medium reasoning effort for the \textit{Criteria Match Validator Agent}; and \textit{gpt-5} with medium search context and medium reasoning effort for both the \textit{Deduplication Agent} and the \textit{Coach Agent}. During the \textit{Rollout} step, we enabled multilingual parallelism by running Investigator Agents in \textit{English} and \textit{Chinese}. In the \textit{Expand} step, the Coach Agent generated ($k=3$) non-overlapping child directives per node, and we used \textit{light deduplication} throughout. For cost optimization purporses we run Bioptic Agent with 1 parallel exploration on the full benchmark, so in fact, there are 10 and Investigator agents executed by the end of the 5th epochs and 20 if Bioptic Agent run for 10 epochs.

\textbf{Bioptic Agent (no-tree, lang-free) ablation}
This ablation preserves Bioptic Agent's components (Coach reflection, validators, and deduplication) but removes tree-structured exploration. At each epoch, the Coach produces a flat set of $k$ directives; all directives are executed in parallel, their artifacts are merged into a single pool, and the merged pool is passed to the next epoch to generate the next batch of directives. We also disable multilingual parallelism (we do not spawn separate Investigator instances per language and we do not constrain queries to a specific language).
We set $k=5$ parallel directives per epoch. Therefore, running for 5 epochs executes 25 Investigator calls, and running for 10 epochs executes 50 Investigator calls.

\begin{table}[!t]
\centering
\footnotesize
\setlength{\tabcolsep}{3pt}
\renewcommand{\arraystretch}{1.12}
\begin{tabularx}{\columnwidth}{@{}>{\raggedright\arraybackslash}X
  S[table-format=1.3] S[table-format=1.3] S[table-format=1.3]@{}}
\toprule
\textbf{Model} & \textbf{Recall} & \textbf{Precision} & \textbf{F1} \\
\midrule
Bioptic Agent (GPT-5.2, high)              & \textbf{0.730} & \textbf{0.877} & \textbf{0.797} \\
Gemini 3.1 Deep Think  & 0.636 & 0.554 & 0.592 \\
Gemini 3.1 Pro Deep Research & 0.545 & 0.634 & 0.586 \\
Claude Opus 4.6 (high) & 0.454 & 0.736 & 0.562 \\
Gemini 3 Pro Deep Research                             & 0.500 & 0.512 & 0.506 \\
OpenAI Deep Research (o4-mini)       & 0.372 & 0.713 & 0.489 \\
GPT-5.2 Pro (high)                        & 0.364 & 0.648 & 0.466 \\
Perplexity Sonar Deep Research (high)                  & 0.409 & 0.481 & 0.442 \\
GPT-5.2 (high)                            & 0.182 & 0.683 & 0.287 \\
Exa Websets (\texttt{num\_matches}=500)            & 0.182 & 0.515 & 0.269 \\
\bottomrule
\end{tabularx}
\caption{\textbf{Assets discovery performance}. Recall/Precision/F1 on the asset scouting eval, sorted by F1 score (all metrics higher are better).}
\label{tab:agent_comparison}
\end{table}

\subsection{Results and Discussion}
Table~\ref{tab:agent_comparison} shows that Bioptic Agent achieves the best overall performance, with an F1-score of $0.797$, substantially outperforming all tested state-of-the-art search agents. The strongest baseline is Gemini 3.1 Deep Think with an F1-score of $0.592$, followed by Gemini 3.1 Pro Deep Research with an F1-score of $0.586$, Claude Opus 4.6 with $0.562$, Gemini 3 Pro Deep Research with $0.506$, \textit{o4-mini-deep-research} with $0.489$, GPT-5.2 Pro high with $0.466$, Perplexity Sonar Deep Research high with $0.442$, GPT-5.2 high with $0.287$, and Exa Websets with $0.269$. Bioptic Agent's advantage is driven by simultaneously high precision of $0.877$ and high recall of $0.730$, whereas competing systems exhibit markedly lower recall and more limiting precision--recall tradeoff.


Figure~\ref{fig:quality_over_time} further indicates that Bioptic Agent improves rapidly in the early regime and then approaches a plateau near $\sim 0.80$ F1-score. Notably, Bioptic Agent uses GPT-5.2 as its underlying model, making it cheaper than having a scaffold on top of Opus 4.6, which is a much heavier model. 


The sequential scaffold ablations with \textit{o4-mini-deep-research} and \textit{gpt-5.2} improve more slowly and plateau earlier at markedly lower F1-score. The sequential scaffold over the \textit{gpt-5.2} (the blue plot in the Figure~\ref{fig:quality_over_time}) uses the same underlying model and the same prompts as our Bioptic Agent Investigator. The results, therefore, indicate that simply iterating to find more assets with a growing list of prior candidates is insufficient for sustained coverage: once obvious sources and search angles are exhausted, the sequential loop tends to revisit similar evidence and exhibits diminishing returns. By comparison, Bioptic Agent's higher and more sustained gains suggest that its scaffold architecture is the primary driver of improved completeness rather than model strength and prolonged executions alone. 

We include OpenAI Deep Research (o4-mini) in the sequential scaffold ablation to test a natural hypothesis suggested by Table~\ref{tab:agent_comparison}: since Gemini 3.1 Pro Deep Research is the second strongest after Bioptic Agent, perhaps simply running a state-of-the-art Deep Research agent longer---by iterating a sequential ``find more'' loop---could close the gap to Bioptic Agent. Directly applying many sequential epochs to Gemini 3 Pro Deep Research is costly, so we use \textit{o4-mini-deep-research} as a cost-efficient alternative that is comparable in terms of F1-score and itself a Deep Research \emph{scaffolded} agent. Empirically, however, the sequentially wrapped \textit{o4-mini-deep-research} still improves more slowly and saturates earlier than Bioptic Agent, similarly to how the \textit{gpt-5.2}-based scaffold is saturated, consistent with diminishing returns once the most obvious sources and search angles are exhausted. This demonstrates that the gains are not in simply running a state-of-the-art general agent longer; rather, Bioptic Agent’s advantage comes from its specialized tree-based exploration, self-reflection, and self-learning scaffolding beyond naive sequential iteration.





Finally, we compare the Bioptic Agent against the Bioptic Agent (no-tree, no-multilingual) ablation. Removing the directive tree and disabling multilingual parallelism yields comparable quality through roughly the 5th epoch, but the ablation saturates thereafter. Importantly, this saturation occurs despite higher compute: at 10 epochs, the no-tree variant executes 50 Investigator calls versus 20 Investigator calls for the full Bioptic Agent setting shown in Figure~\ref{fig:quality_over_time}. This suggests that (i) tree-based exploration helps prevent early saturation by systematically allocating compute to under-explored branches, and (ii) multilingual rollout provides additional coverage on locally announced and non-English-first assets. 
This experiment also shows how much of the performance gain comes from self-reflection and self-learning, implemented via the internal \textit{evaluation step}, the Coach Agent’s use of search history and error and gap analysis, and the Investigator Agent’s automatic prompt refinement with parallel execution of conditioned directives, compared to vanilla sequential GPT-5.2 execution.

\section{Conclusion}
In this paper we present (i) a challenging Completeness Benchmark for drug asset scouting and (ii) Bioptic Agent, a tree-based, multilingual wide-research system optimized for complete, non-hallucinated “find-all” scouting. The benchmark is constructed backward from validated program records, mined predominantly from non-U.S., non-English ecosystems, and paired with investor-native, multi-constraint screening queries. 


On this benchmark, Bioptic Agent achieves the best overall performance with F1-score of 79.7\% and substantially outperforms state-of-the-art commercial deep-research baselines (Table 2), including Claude Opus 4.6 high (56.2\%), Gemini 3.1 Deep Think (59.2\%), Gemini 3.1 Pro Deep Research (58.6\%), Gemini 3 Pro Deep Research (50.6\%), o4-mini-deep-research (48.9\%), GPT-5.2 Pro high (46.6\%), Perplexity Sonar Deep Research high (44.2\%), and Exa Websets (26.9\%). Figure~\ref{fig:quality_over_time} shows a clear quality–time tradeoff: Bioptic Agent improves rapidly early and then approaches a plateau near $\sim0.80$ F1, while sequential “run longer” scaffolds improve more slowly and plateau earlier at lower quality.




\bibliography{aaai2026}

\end{document}